\documentclass[conference]{IEEEtran}
\usepackage{times}

\usepackage[numbers]{natbib}
\usepackage{multicol}
\usepackage[bookmarks=true]{hyperref}
\usepackage{amsmath}
\usepackage{amsfonts}
\usepackage{mathtools}
\usepackage{algorithm}
\usepackage[noend]{algpseudocode}
\usepackage{subcaption}
\usepackage{bm}
\usepackage{balance}

\usepackage{graphicx}
\graphicspath{{figures/}{figures/example_trajs/}}

\DeclareMathOperator*{\argmin}{arg\,min}

\begin{document}

\title{Motion Prediction with Recurrent Neural Network  Dynamical Models and Trajectory Optimization}

\author{Philipp Kratzer$^{1}$, Marc Toussaint$^{1}$ and Jim Mainprice$^{1,2}$\\
\vspace{0.1cm}
\authorblockA{\tt{\small{firstname.lastname@ipvs.uni-stuttgart.de}}}
\authorblockA{$^1$Machine Learning and Robotics Lab, University of Stuttgart, Germany}
\authorblockA{$^2$Max Planck Institute for Intelligent Systems ;  IS-MPI ; T{\"u}bingen, Germany}
\vspace{-0.8cm}
}



%

\maketitle

\begin{abstract}
Predicting human motion in unstructured and dynamic environments is difficult as humans naturally exhibit complex behaviors that can change drastically from one environment to the next. In order to alleviate this issue, we propose to encode the lower level aspects of human motion separately from the higher level geometrical aspects, which we believe will generalize better over environments. In contrast to our prior work~\cite{kratzer2018}, we encode the short-term behavior by using a state-of-the-art recurrent neural network structure instead of a Gaussian process.
In order to perform longer term behavior predictions that account for variation in tasks and environments, we propose to make use of  gradient based  trajectory optimization.
Preliminary experiments on real motion data demonstrate the efficacy of the approach.
\end{abstract}

\IEEEpeerreviewmaketitle

\section{Introduction}
\let\thefootnote\relax\footnotetext{
Workshop on
``AI and Its Alternatives in Assistive and Collaborative Robotics" (RSS 2019), Robotics: Science and Systems Freiburg, Germany.}
As robots become more capable they will inevitably share the workspace with humans. 
This close proximity between humans and robots poses a certain number of challenges
ranging from the robots design to the algorithms involved in controlling them \cite{broquere2014attentional}.
In this paper, we tackle the ability for robots to model human behavior in order to
anticipate the surrounding humans movement. 
This capability is especially useful to execute a shared human-robot task or
even to mimic humans.

Predicting motion in unstructured and dynamic environments is difficult as humans 
exhibit complex behaviors that can change drastically from one environment to the next.

In our prior work~\cite{kratzer2018} we have proposed to learn the 
lower level aspects of human motion separately, which we believe generalize over environments and account for environmental constraints in a later trajectory optimization step.
We modeled the dynamics of the human using a Gaussian Process (GP), which abstracts all phenomena linked to complex bio-mechanical processes to produce purely
kinematic predictions. In order to account for the context and produce a longer horizon prediction, we optimized the prediction of the GP together with the higher level constraints.

This technique has several advantages, 
1) decoupling learning of the dynamics holds the promise to generalize better than learning
all level of abstractions in one policy,
2) the implementation is simpler than incorporating Newtonian dynamics,
3) modularity of the model (dynamics/kinematics) makes retargeting behavior straight forward.

However, the main disadvantage of the method is its limited scalability due the nature of the GP, requiring to compare against all training points in the data set to predict the next state. Especially for motion prediction, one can capture a big amount of motion data using motion capture systems, which makes the use of GPs intractable.

Recent work on motion prediction focuses on recurrent neural networks~\cite{fragkiadaki2015recurrent, martinez2017human, pavllo2018quaternet}. Martinez et al.~\cite{martinez2017human} reported on using a sequence-to-sequence architecture that outperformed prior RNN based methods. While the results show impressive performance for short-term motion prediction, it is not suitable for long term predictions as errors accumulate and the predictions either tend to converge towards an average pose or become noisy.

In this paper we adapt the model by Martinez et al.~\cite{martinez2017human} and show that it can be used in a later trajectory optimization step to improve the predicted trajectory by incorporating environmental constraints, similar to our prior work~\cite{kratzer2018}. The framework allows to account for external constraints during movement that may arise form the context (environment or task), such as
obstacles or orientation of held object, here we simply treat goal set constraints. Note that integrating other constraints would be straightforward. We show preliminary results on experiments with reaching motion. In contrast to our prior work we predict full-body reaching motions.
\begin{figure}
\centering
\includegraphics[width=\linewidth]{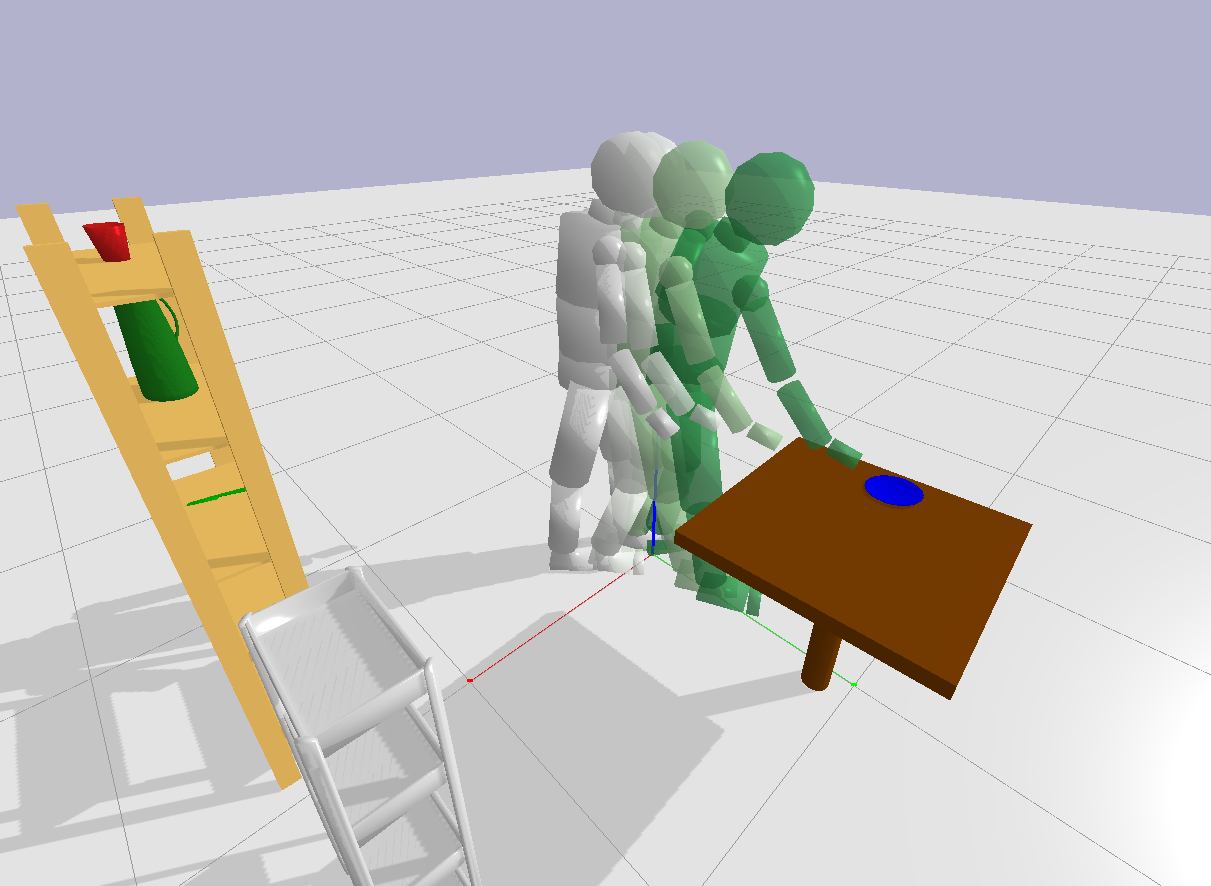}
\caption{Prediction of human motion towards a plate (blue) by our method}
\label{fig:pred_plate}
\end{figure}
\section{Related Work}
Prior work has made use of graphical models, such as Hidden Markov Models (HMMs) or Conditional Random Fields (CRFs), to predict human motion. Lehrmann et al. used HMMs to retain a dynamic model of human motion and reported good results for motion completion tasks~\cite{lehrmann2014efficient}.
In~\cite{koppula2016anticipating}, Koppula and Saxena predicted trajectories of the human hand using CRFs.  Their approach samples possible trajectories by taking object affordances into account. 
However, while graphical approaches capture relationships between objects well they do not allow for additional constraints that may come from the environment, an issue that we address in this paper.

Another approach for predicting human motion is Inverse Optimal Control (IOC) which aims to find a cost function underlying the observed behavior.
In~\cite{berret2011evidence}, Berret et al. investigated cost functions for arm movement planning and report that such movements are closely linked to the combination of two costs related to mechanical energy expenditure and joint-level smoothness.
In \cite{mainprice2016goal}, Mainprice et al. investigated prediction of human reaching motions in shared workspaces. Using goal-set IOC and iterative replanning, the proposed method accounts for the presence of a moving collaborator and obstacles in the environment using a stochastic trajectory optimizer. IOC methods typically represent bio-kinematic processes by simplified models, which not necessarily generalize well. In contrast our method learns these processes from data using a recurrent neural network.  

Recent work on human motion prediction for short-term motion has focused on neural network architectures~\cite{fragkiadaki2015recurrent, li2018convolutional, martinez2017human, pavllo2018quaternet}. Recurrent Neural Networks (RNNs) have been used, because of their ability to store relevant recurrent information in the hidden state. For example, Fragkiadaki et al. proposed a RNN based model that incorporates nonlinear encoder and decoder networks before and after recurrent layers~\cite{fragkiadaki2015recurrent}. Their model is able to handle training across multiple subjects and activity domains. With a similar approach Martinez et al.~\cite{martinez2017human} reported on using a sequence-to-sequence architecture that outperformed prior RNN based methods.

Our work differs in several aspects from prior work in human motion prediction. 
First, our approach remains low in complexity by not relying on a bio-mechanical model,
instead encoding the short-term behavior in a data driven dynamical system. Second, we account for additional constraints by optimizing the predicted trajectory with respect to a cost function. This makes it possible to handle environmental constraints, such as the distance to target states.

\section{Method}
\label{sec:method}
\begin{figure}
\centering
\includegraphics[width=\linewidth]{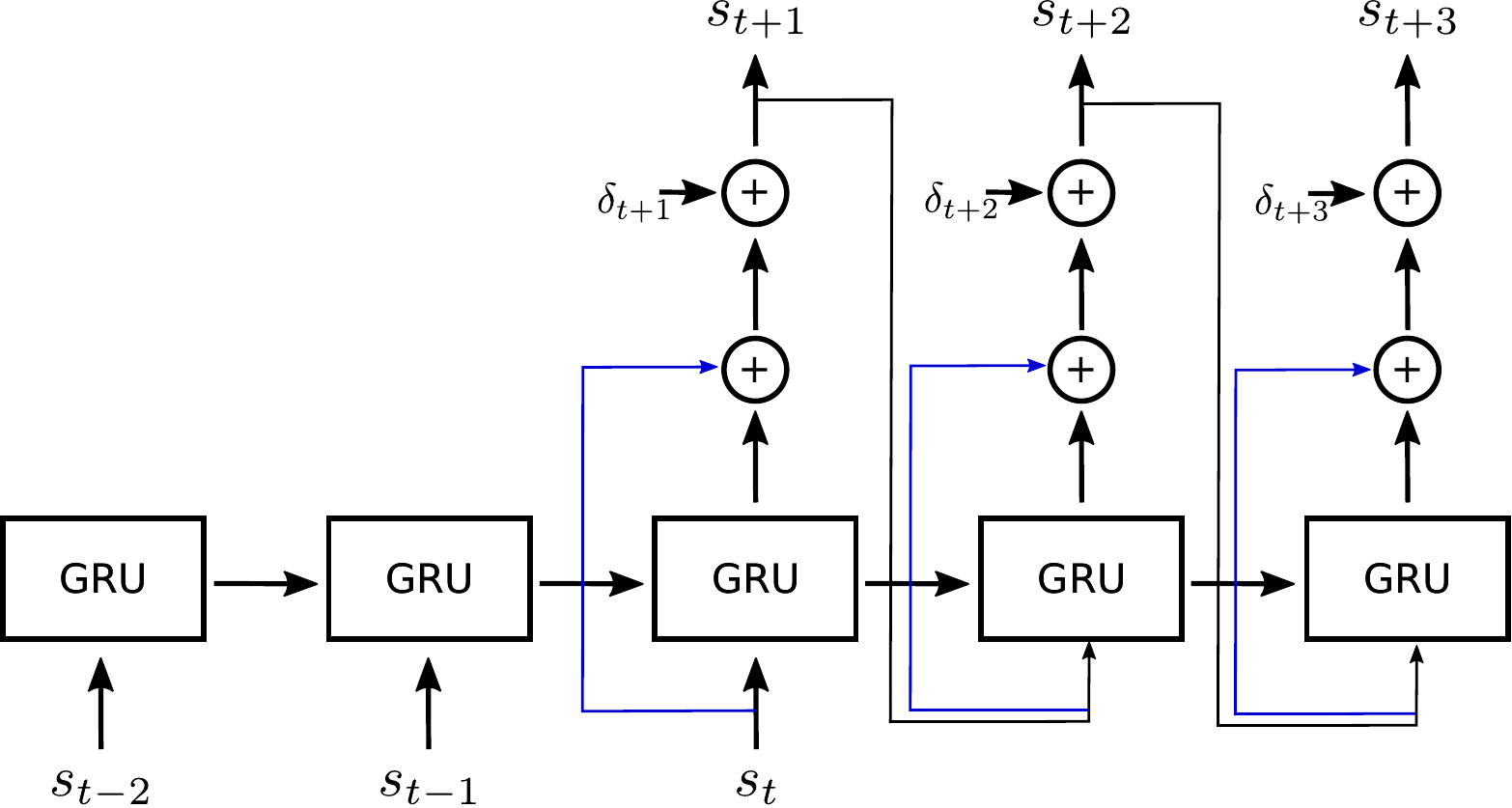}
\caption{Sequence to sequence architecture by Martinez et al.~\cite{martinez2017human} with additional $\delta$ to change prediction states during motion optimization.}
\vspace{-.7cm}
\label{fig:architecture}
\end{figure}
Our approach works in two phases: 1) offline we learn a predictive model of the human $s_{t+1:T} = f(s_{0:t})$ where $s_{0:t}$ is the observed trajectory of human states. The aim of $f$ is to predict the kinematic states of the human in the next time steps up to a prediction horizon $T$, based on a sequence of previous states. This is achieved by supervised training of a recurrent neural network model on human motion capture data, 2) online we use the learned model to predict future states. The prediction is optimized to fulfill constraints by varying the predictions $s_{t+1:T}$ at every timestep.

\subsection{Recurrent Neural Network}
For predicting a motion trajectory $s_{t+1:T}$,
we use the recurrent neural network model by Martinez et al.~\cite{martinez2017human}.
The model is shown in Figure~(\ref{fig:architecture}). It consists of Gated Recurrent Units (GRU).
The model has a residual connection (depicted blue) at the output GRUs in order to make the model predict the change of state instead of the state itself which improves prediction performance.

The model is trained on data $\mathcal{D}=(s_{i, 0:T})_i^N$ with $N$ being the number of demonstrated trajectories. The trajectories are split into smaller trajectories $s_{i, 0:t}$ with labels $s_{i, t+1:T}$.
The states are represented by the base translation of the human and the joint angles in exponential map representation. For training we follow~\cite{pavllo2018quaternet} and use a $2 \pi$ wrap around loss in the Euclidean angle representation to avoid wrap around ambiguities.
\subsection{Optimizing the Trajectory}
At test time the recurrent neural network model can be used to predict the future trajectory $s_{t+1:T}$.
To further improve the predicted trajectory we wish to account for environmental constraints
explicitly.
Thus we want to find a trajectory $s_{t+1:T}^*$ that maximizes the likelihood
of the next states given our data, fulfills the human dynamics $s_{t+1} = d(s_t)$ and
satisfies environmental constraints $h$ and $g$ :
\begin{align}
  \max \sum_t^T p(s_{t+1:T}^*|s_{0:t}, \mathcal{D})&\\
    \text{subject to } s_{t+1} &= d(s_t) \notag\\
    h(s_{t+1:T}^*) &= 0 \notag\\
    g(s_{t+1:T}^*) &\leq 0 \notag
\end{align}

The most likely next states based on the training data $\mathcal{D}$ 
as well as the dynamic constrained $s_{t+1} = d(s_t)$
are approximated by the recurrent neural network model $s_{t+1:T} = f(s_{0:t})$.
In order to improve the predicted trajectory $s_{t+1:T}$,
we slightly change the predicted states to fulfill the constraints. 
Hence, we add $\delta_{t+1:T}$ to each output of the network $s_{t+1:T}$ (see Figure \ref{fig:architecture}),
which act as additional input parameters.

During training these additional connections are kept to zero,
but at test time $\delta$ is used to change the states in order to fulfill the constraints.
Note that the $\delta_{t}$ is fed into the GRU for the next time step and
thus changes the predictions for all the following time steps.
Keeping $\delta$ small ensures that the prediction still grounds
on the neural network model
and only slightly guides it towards fulfilling other cost objectives.
Cost objectives can, for example, promote close distances to possible
target states or penalize close distances to obstacles (here we focus on goal set regions).

Hence our prediction algorithm finds an optimal $\delta^*$,
such that the $\delta$ is close to zero
while the position of the hand at the final state $s_{t+1:T}$ is close to the target position $p^*$.
Thus we optimize the unconstrained cost function V($\delta$) :
\begin{equation}
  \label{equ:cost}
  \begin{multlined}
  \delta^* = \underset{\delta}{\argmin}~V(\delta) \\
  V(\delta) = \|\delta\|^2 + \lambda\|\phi_{FK}(f(\delta)_T) - p^*\|^2,
\end {multlined}
\end{equation} 
where $\phi_{FK} : s \mapsto p$ is the forward kinematics of the human,
for example calculating the hand position $p \in \mathbb{R}^3$,
and $\lambda$ a Lagrange multiplier.
Thus $\phi_{FK}(f(\delta)_T)$ computes the hand position on the last predicted time step of the network. 
Note that the recurrent network $f(\delta_{t+1:T})$ is only conditioned on $\delta$
because $s_{0:t}$ is already observed and will not change during prediction.

The gradient of V is given by:
\begin{equation}
  \nabla V(\delta) = 2\delta + 2\lambda(\phi_{FK}(f(\delta)_T) - p^*)J_\phi J_f,
\end{equation}
where $J_\phi$ is the Jacobian of $\phi_{FK}$ and
$J_f$ is the Jacobian of the recurrent neural network with respect to $\delta_{t+1:T}$.
We optimize the trajectory using a limited memory version of the numerical optimization algorithm BFGS~\cite{byrd1995limited}. After we computed $\delta^*$ the predicted future trajectory is given by $s_{t+1:T}^* = f(s_{0:t}, \delta_{t+1:T}^*)$.

\begin{figure*}[h!]
\centering
\newcommand{\ltscale}{.185}
\newcommand{\hsscale}{.1cm}
\newcommand{\vsscale}{.2cm}
\begin{subfigure}{\ltscale\textwidth}
\centering
\includegraphics[width=\linewidth]{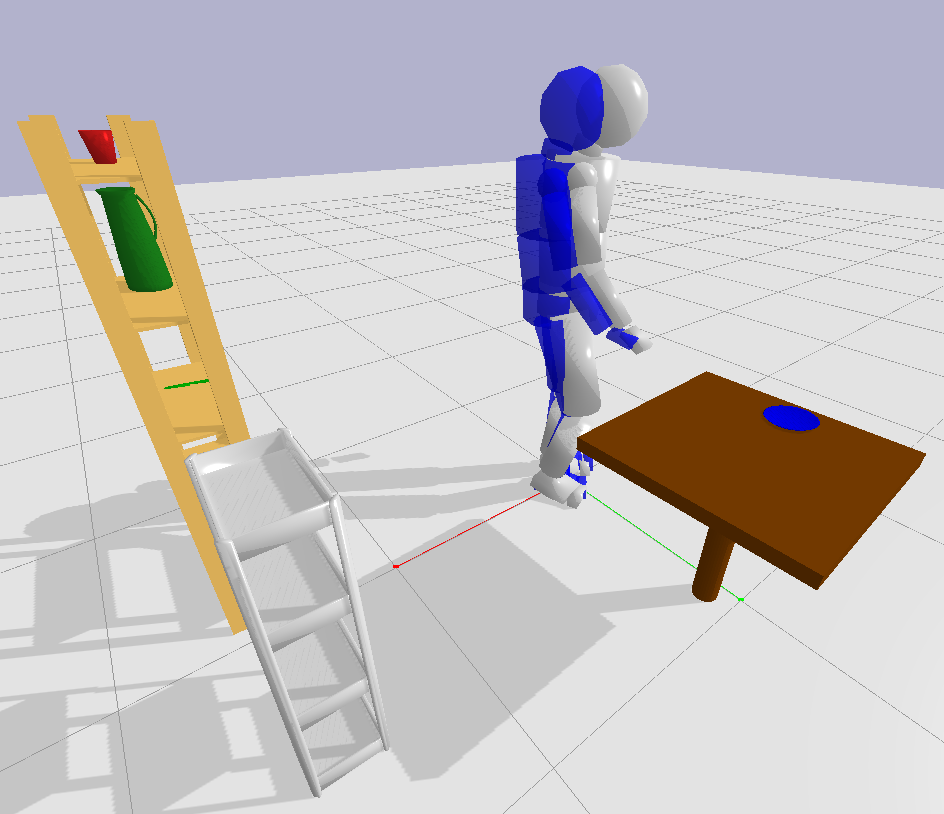}
\end{subfigure}
\hspace{\hsscale}
\begin{subfigure}{\ltscale\textwidth}
\centering
\includegraphics[width=\linewidth]{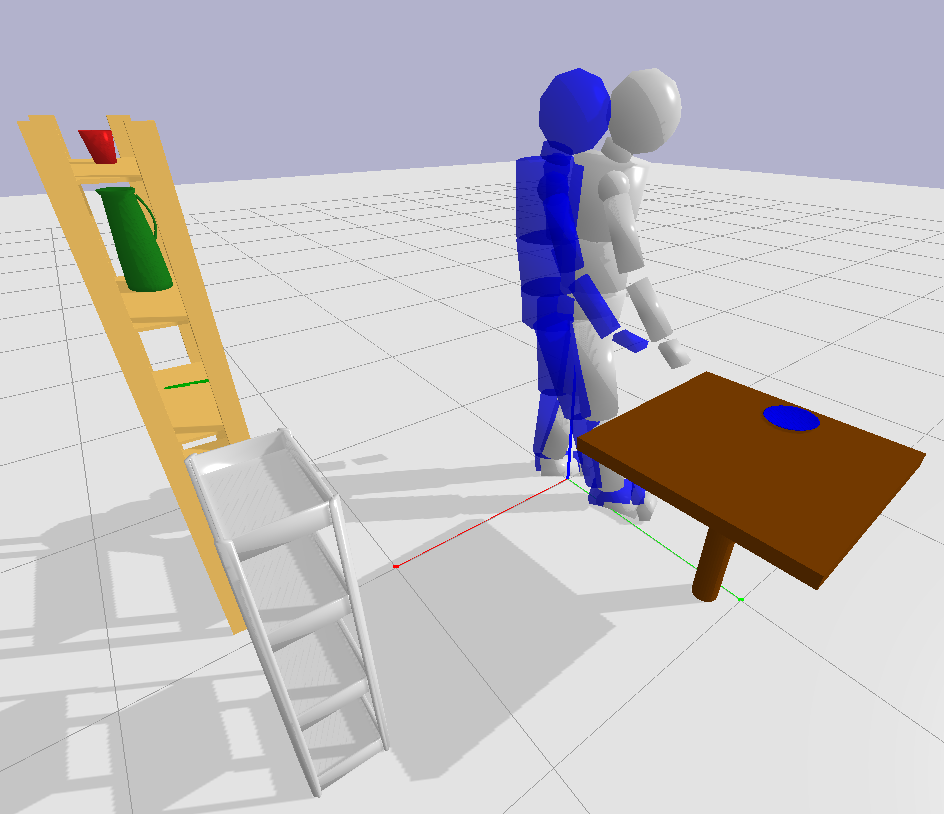}
\end{subfigure}
\hspace{\hsscale}
\begin{subfigure}{\ltscale\textwidth}
\centering
\includegraphics[width=\linewidth]{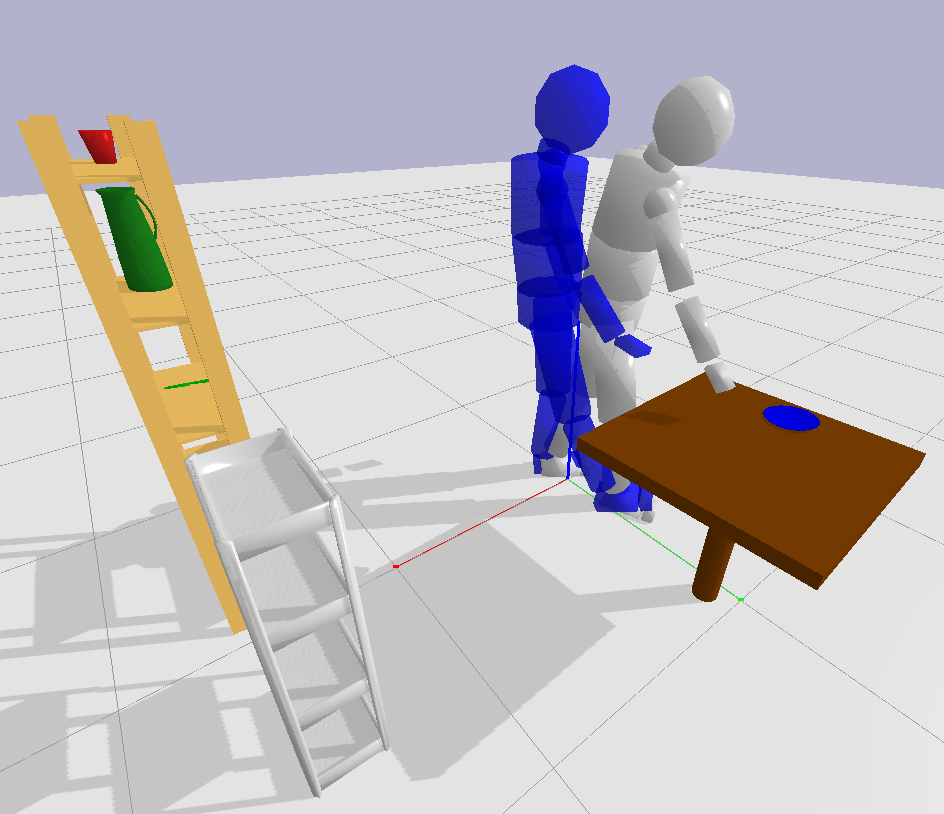}
\end{subfigure}
\hspace{\hsscale}
\begin{subfigure}{\ltscale\textwidth}
\centering
\includegraphics[width=\linewidth]{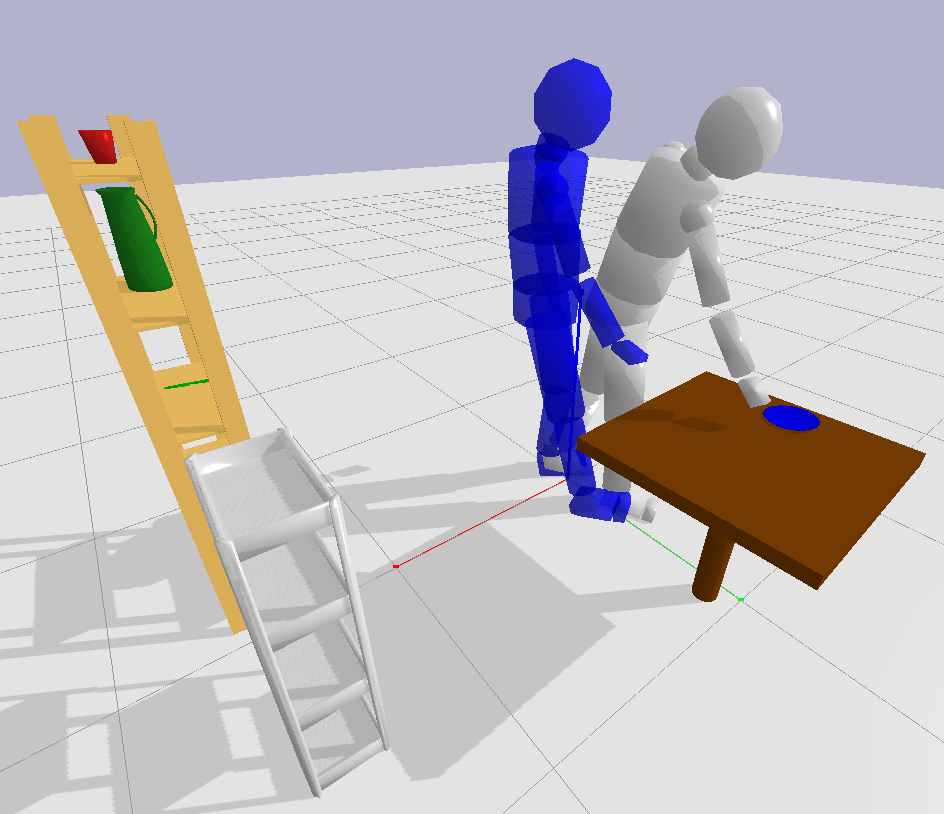}
\end{subfigure}
\hspace{\hsscale}
\begin{subfigure}{\ltscale\textwidth}
\centering
\includegraphics[width=\linewidth]{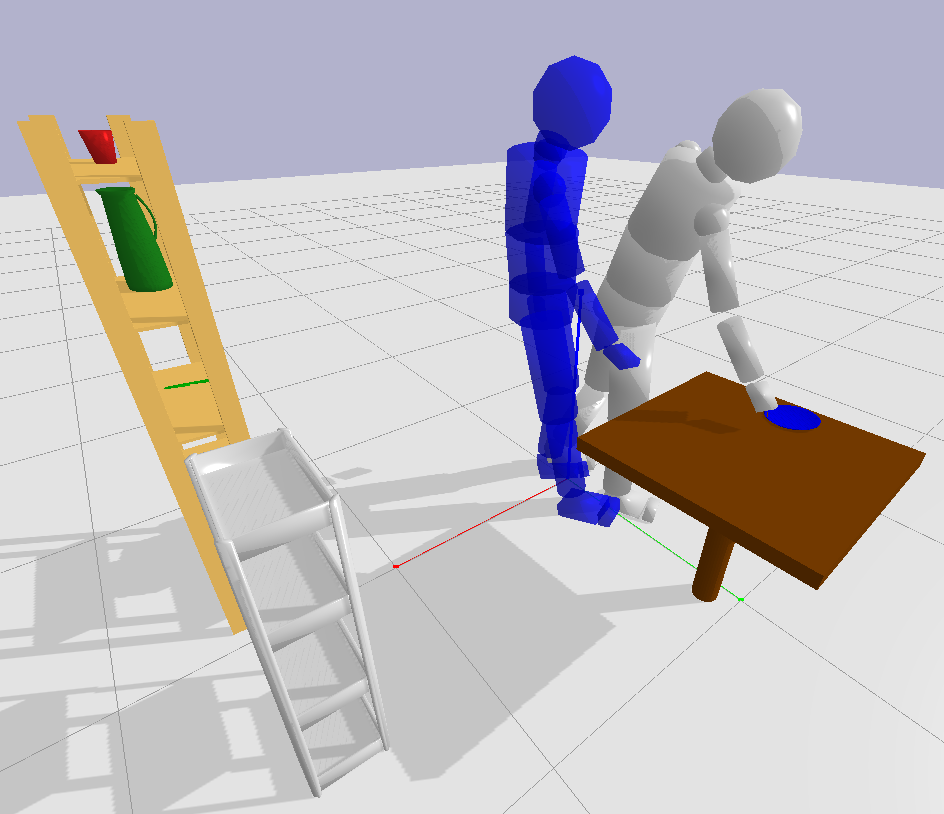}
\end{subfigure}
\\
\begin{subfigure}{\ltscale\textwidth}
\centering
\includegraphics[width=\linewidth]{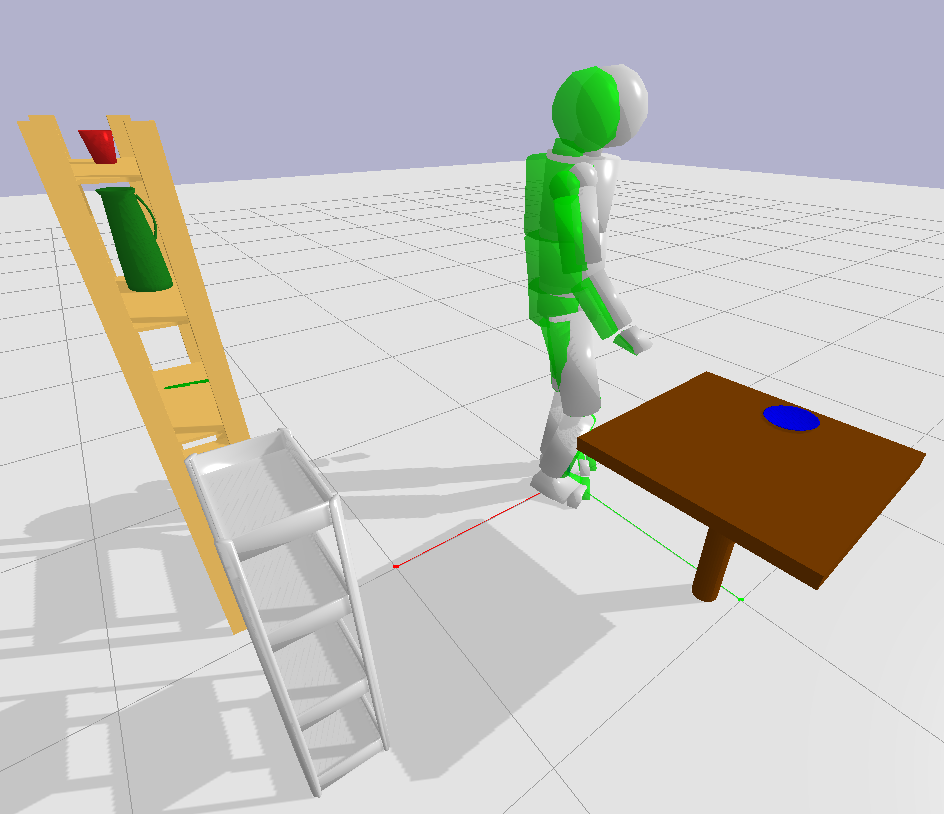}
\end{subfigure}
\hspace{\hsscale}
\begin{subfigure}{\ltscale\textwidth}
\centering
\includegraphics[width=\linewidth]{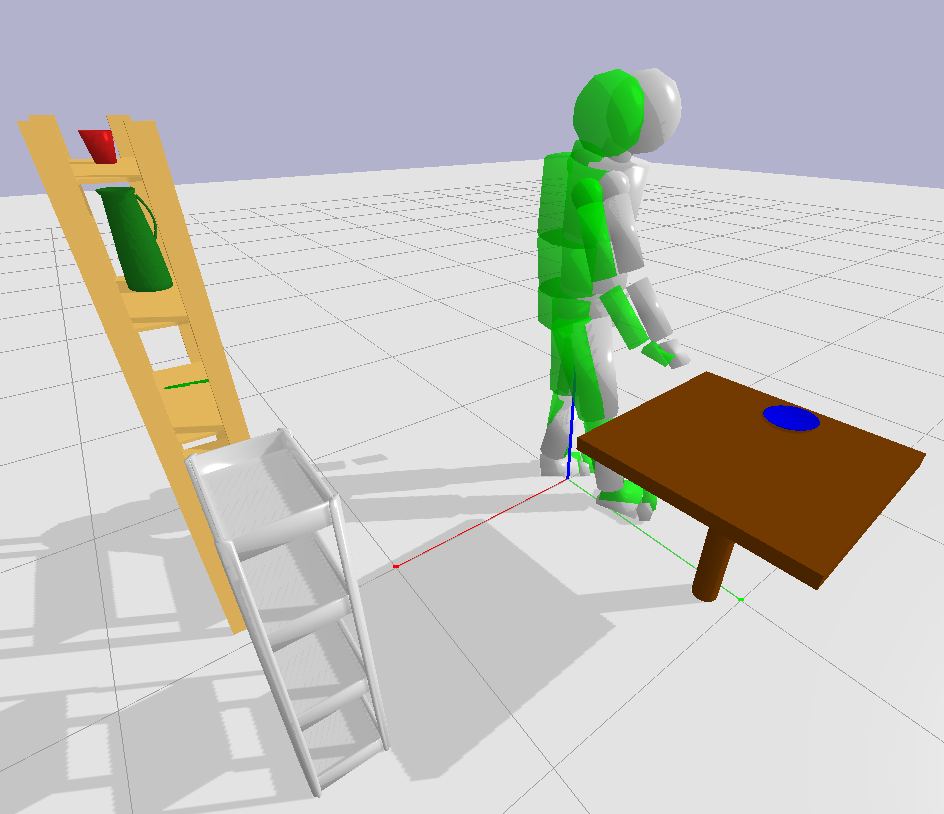}
\end{subfigure}
\hspace{\hsscale}
\begin{subfigure}{\ltscale\textwidth}
\centering
\includegraphics[width=\linewidth]{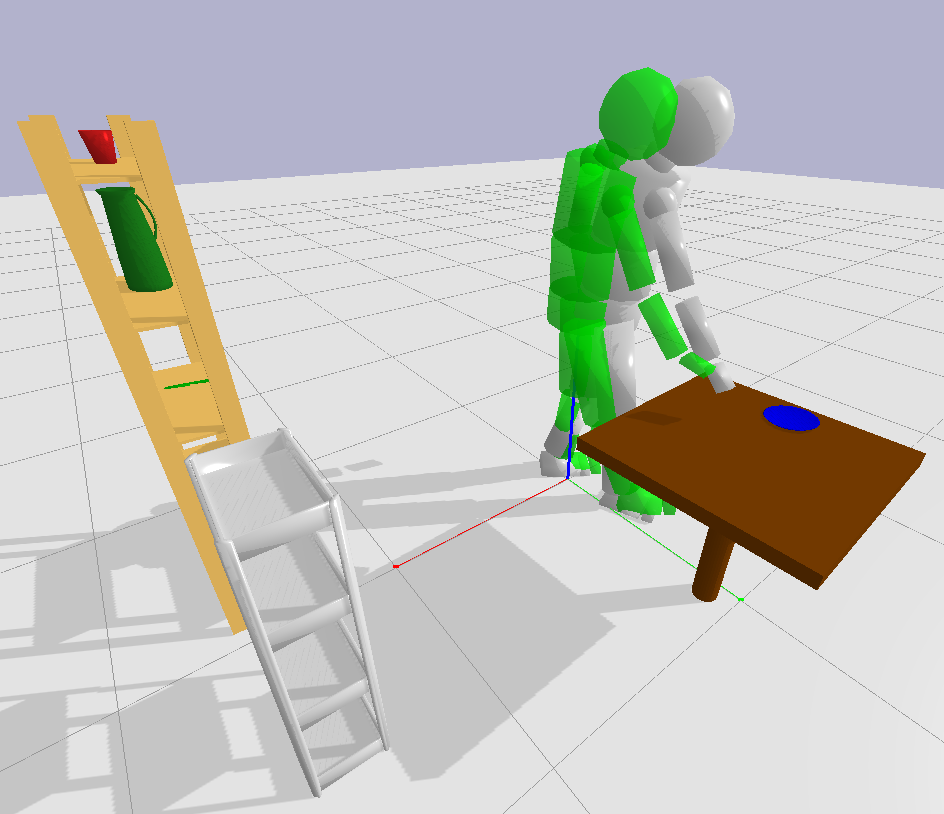}
\end{subfigure}
\hspace{\hsscale}
\begin{subfigure}{\ltscale\textwidth}
\centering
\includegraphics[width=\linewidth]{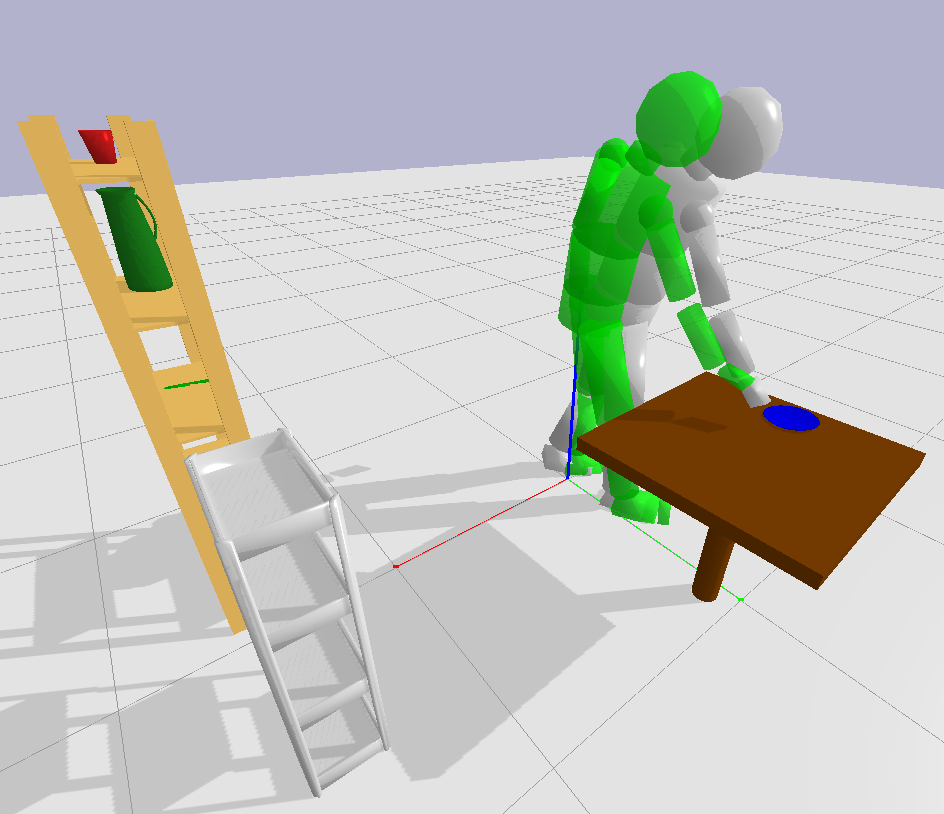}
\end{subfigure}
\hspace{\hsscale}
\begin{subfigure}{\ltscale\textwidth}
\centering
\includegraphics[width=\linewidth]{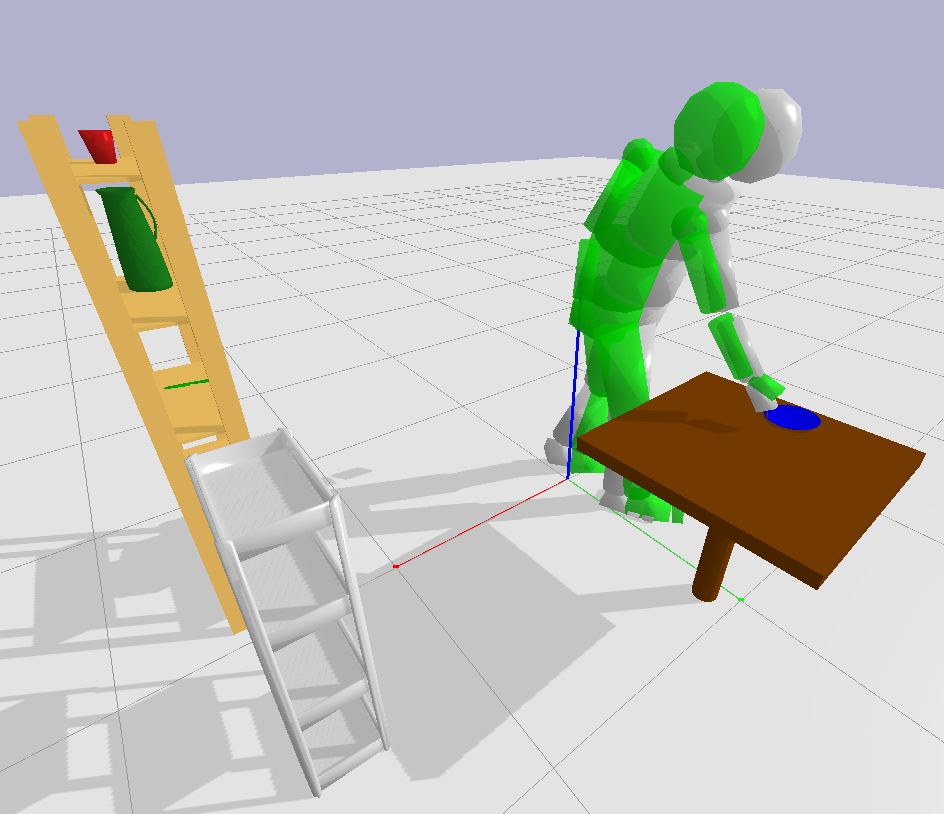}
\end{subfigure}
\\
\vspace{\vsscale}
\centering
\begin{subfigure}{\ltscale\textwidth}
\centering
\includegraphics[width=\linewidth]{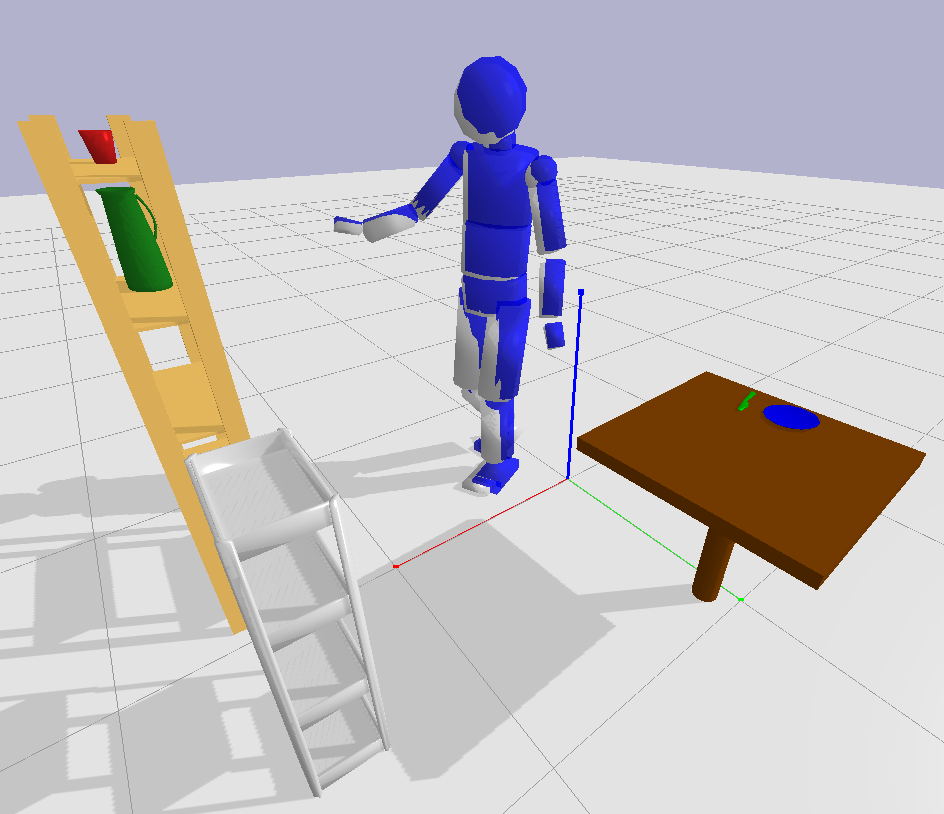}
\end{subfigure}
\hspace{\hsscale}
\begin{subfigure}{\ltscale\textwidth}
\centering
\includegraphics[width=\linewidth]{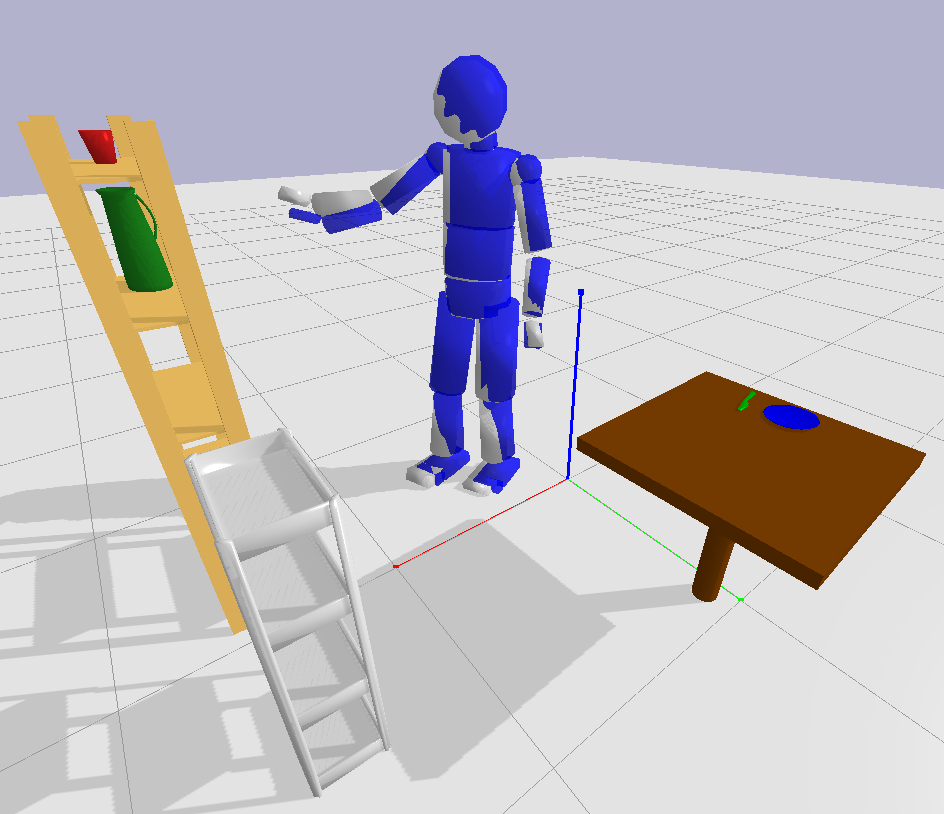}
\end{subfigure}
\hspace{\hsscale}
\begin{subfigure}{\ltscale\textwidth}
\centering
\includegraphics[width=\linewidth]{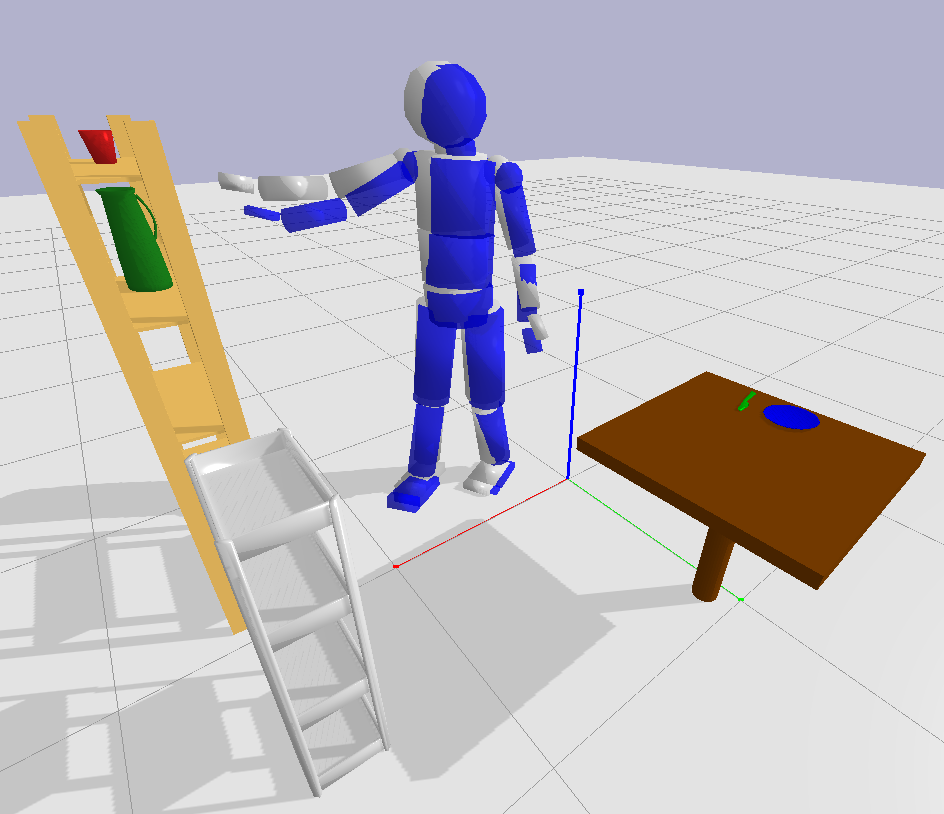}
\end{subfigure}
\hspace{\hsscale}
\begin{subfigure}{\ltscale\textwidth}
\centering
\includegraphics[width=\linewidth]{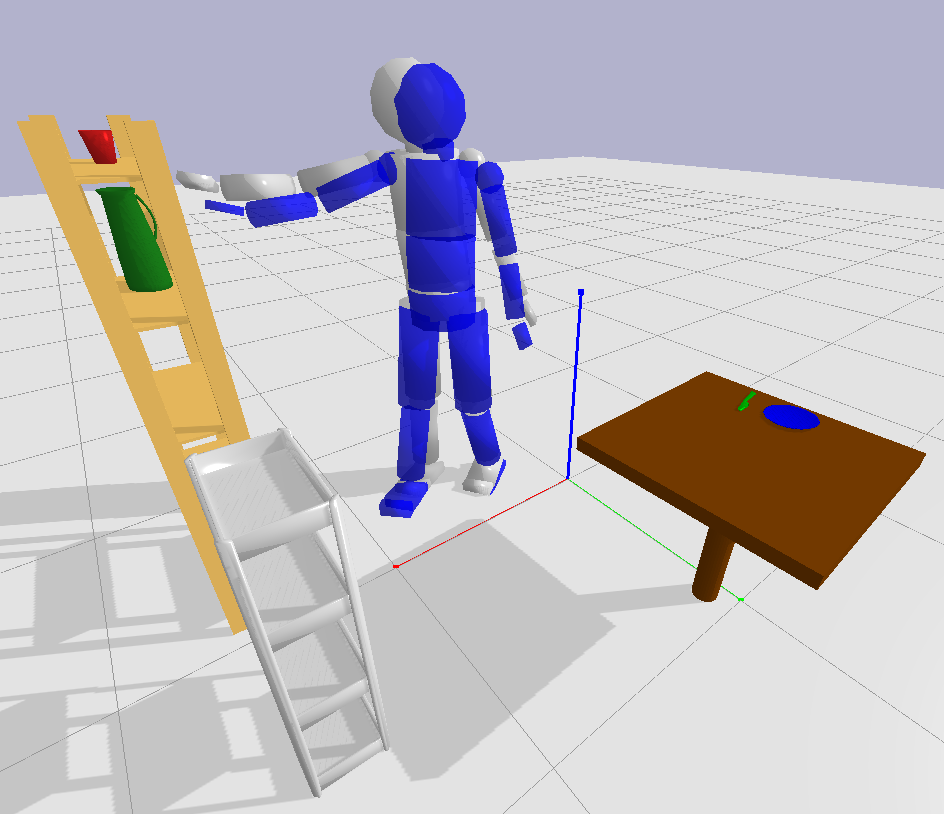}
\end{subfigure}
\hspace{\hsscale}
\begin{subfigure}{\ltscale\textwidth}
\centering
\includegraphics[width=\linewidth]{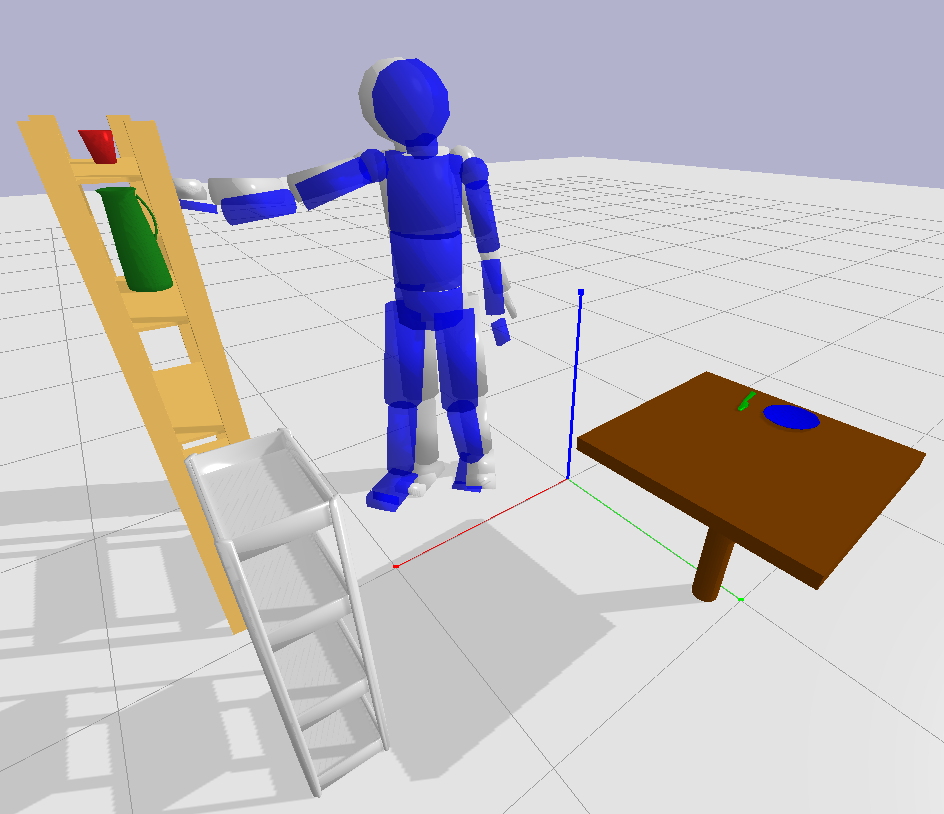}
\end{subfigure}
\\
\centering
\begin{subfigure}{\ltscale\textwidth}
\centering
\includegraphics[width=\linewidth]{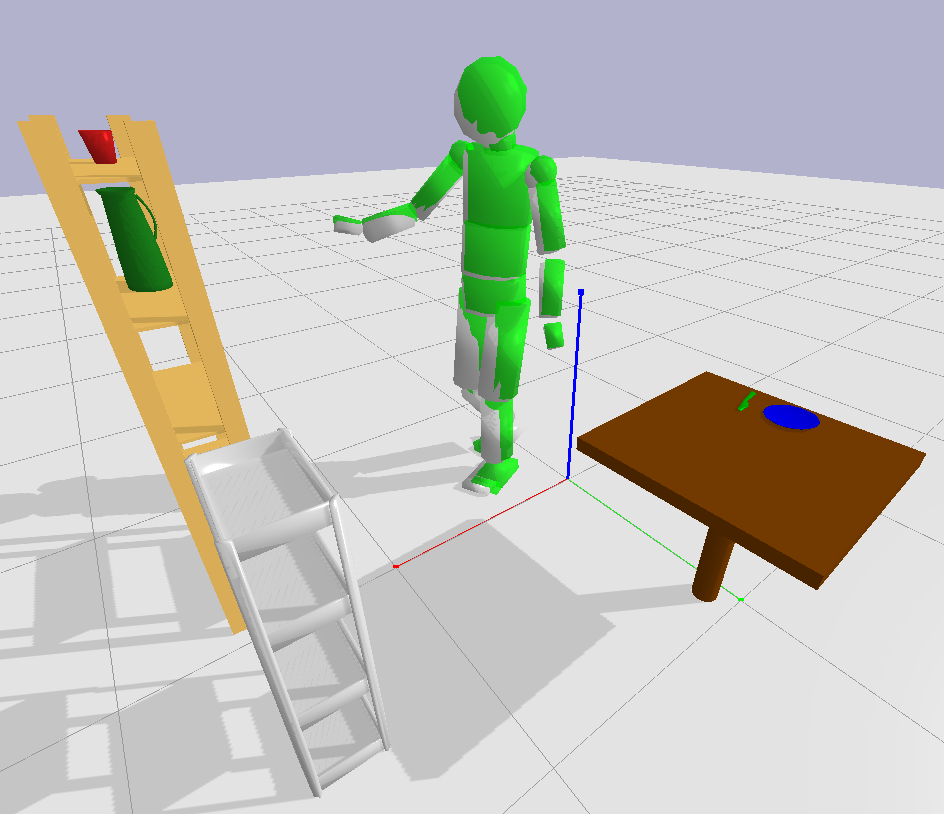}
\end{subfigure}
\hspace{\hsscale}
\begin{subfigure}{\ltscale\textwidth}
\centering
\includegraphics[width=\linewidth]{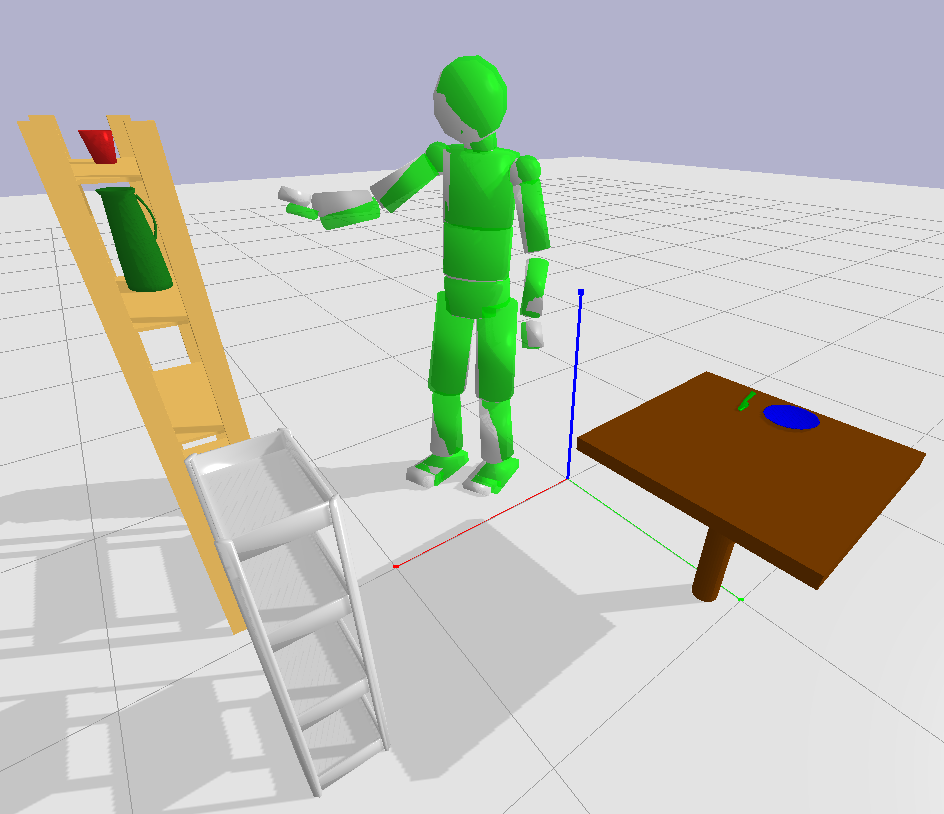}
\end{subfigure}
\hspace{\hsscale}
\begin{subfigure}{\ltscale\textwidth}
\centering
\includegraphics[width=\linewidth]{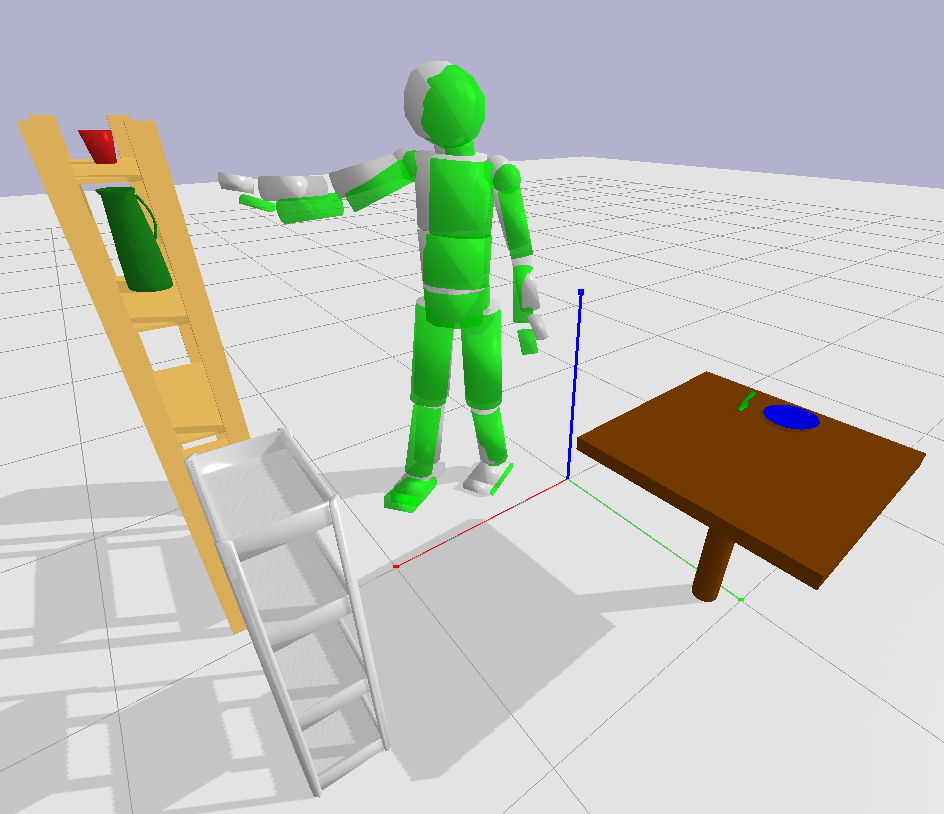}
\end{subfigure}
\hspace{\hsscale}
\begin{subfigure}{\ltscale\textwidth}
\centering
\includegraphics[width=\linewidth]{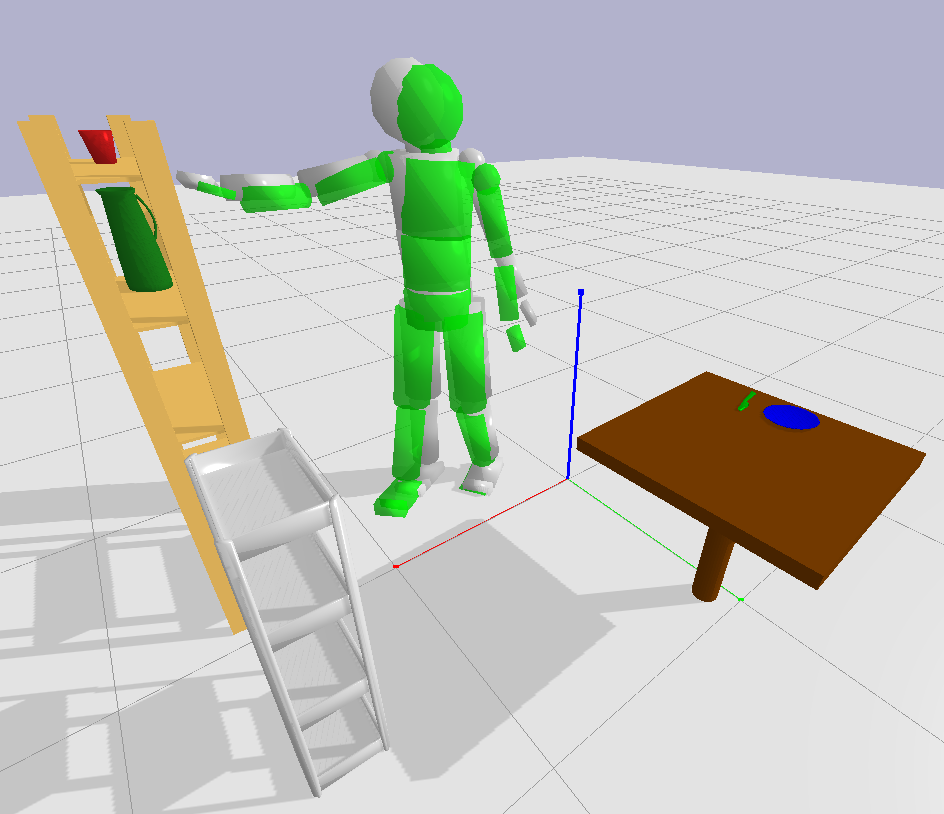}
\end{subfigure}
\hspace{\hsscale}
\begin{subfigure}{\ltscale\textwidth}
\centering
\includegraphics[width=\linewidth]{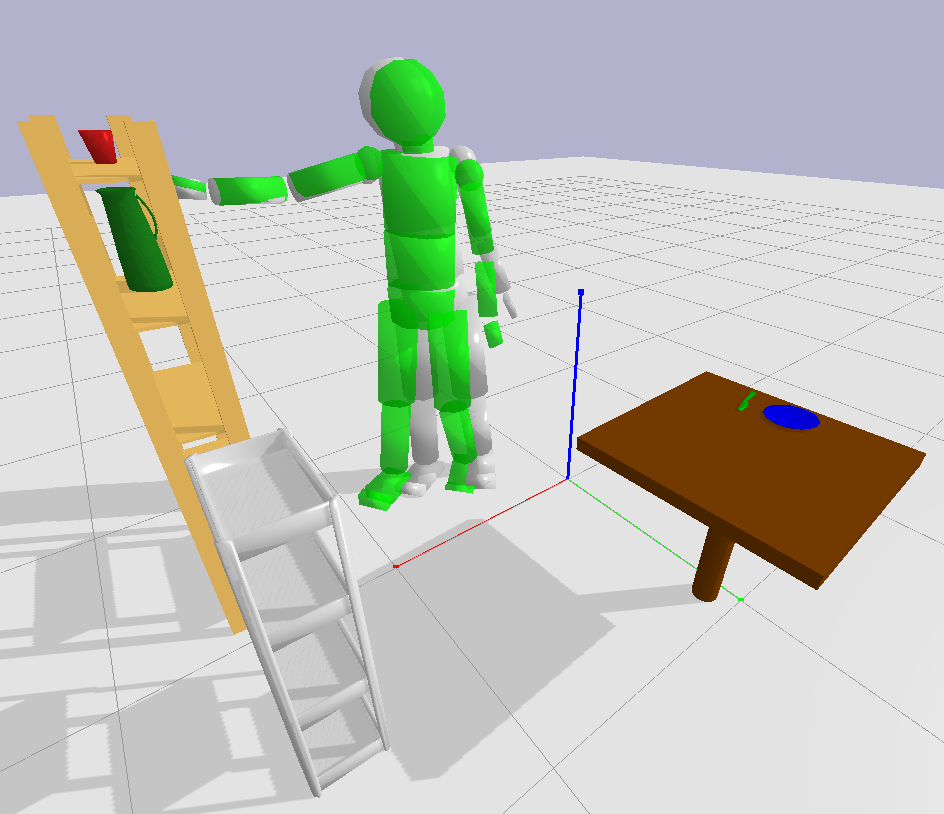}
\end{subfigure}
\\
\vspace{\vsscale}
\centering
\begin{subfigure}{\ltscale\textwidth}
\centering
\includegraphics[width=\linewidth]{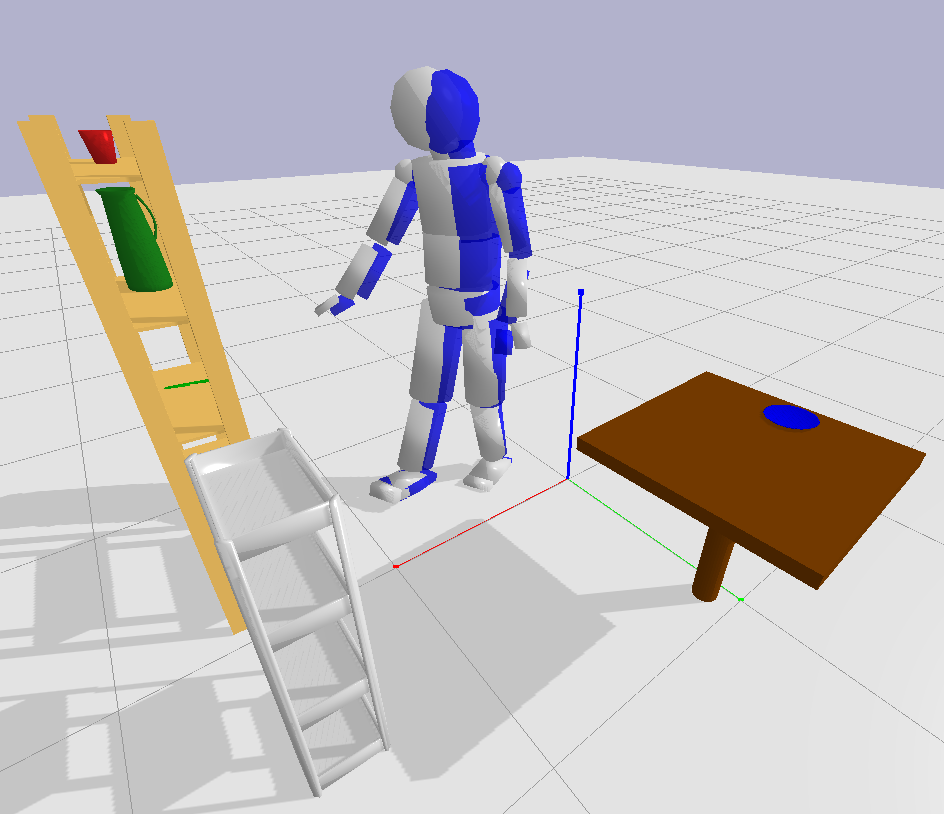}
\end{subfigure}
\hspace{\hsscale}
\begin{subfigure}{\ltscale\textwidth}
\centering
\includegraphics[width=\linewidth]{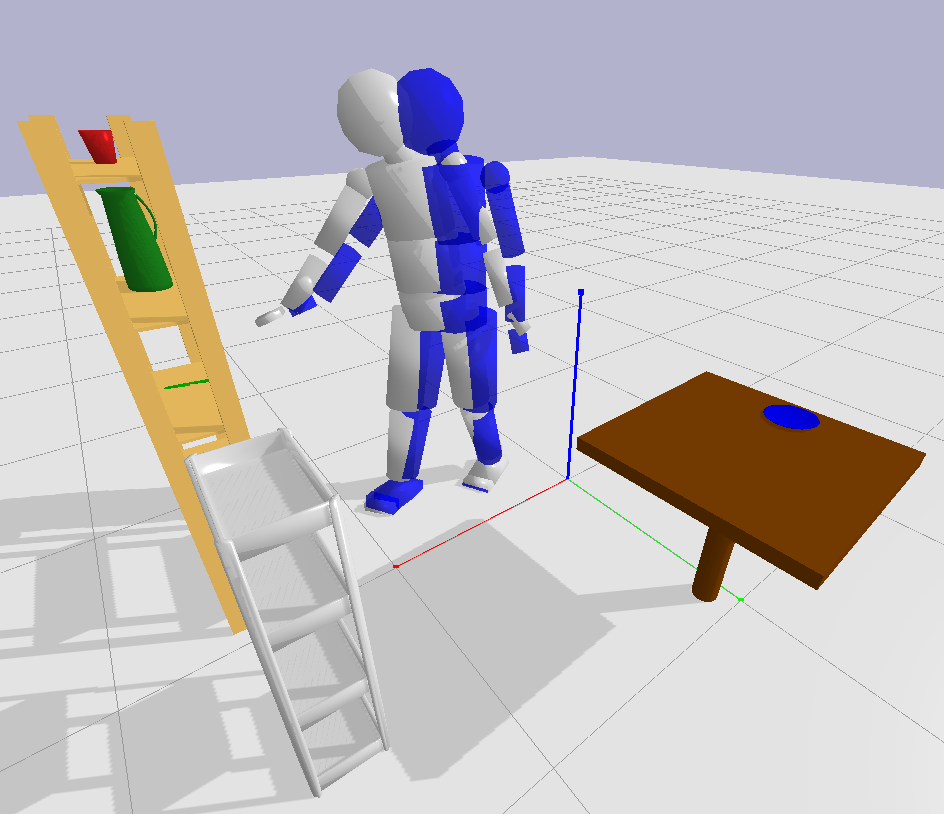}
\end{subfigure}
\hspace{\hsscale}
\begin{subfigure}{\ltscale\textwidth}
\centering
\includegraphics[width=\linewidth]{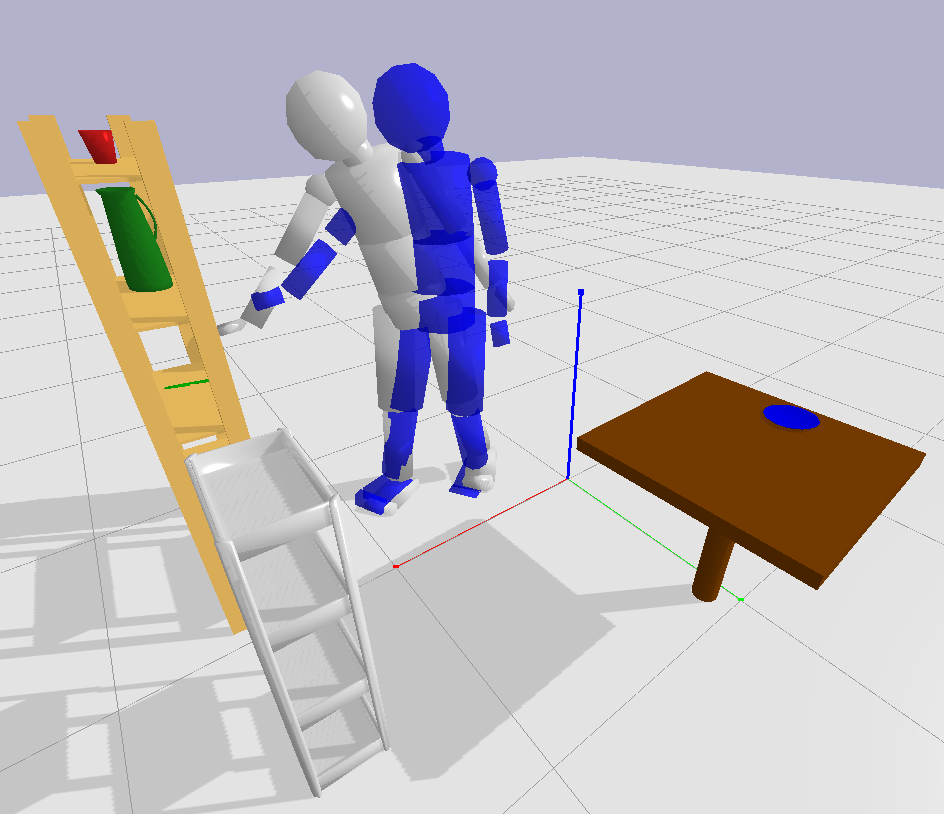}
\end{subfigure}
\hspace{\hsscale}
\begin{subfigure}{\ltscale\textwidth}
\centering
\includegraphics[width=\linewidth]{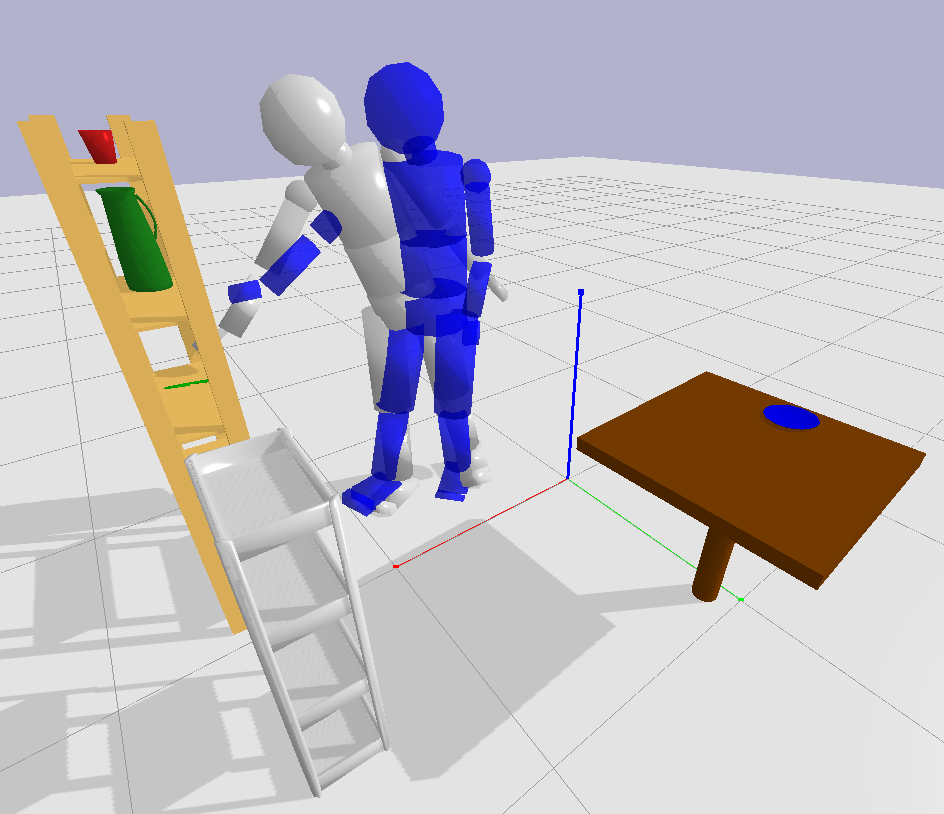}
\end{subfigure}
\hspace{\hsscale}
\begin{subfigure}{\ltscale\textwidth}
\centering
\includegraphics[width=\linewidth]{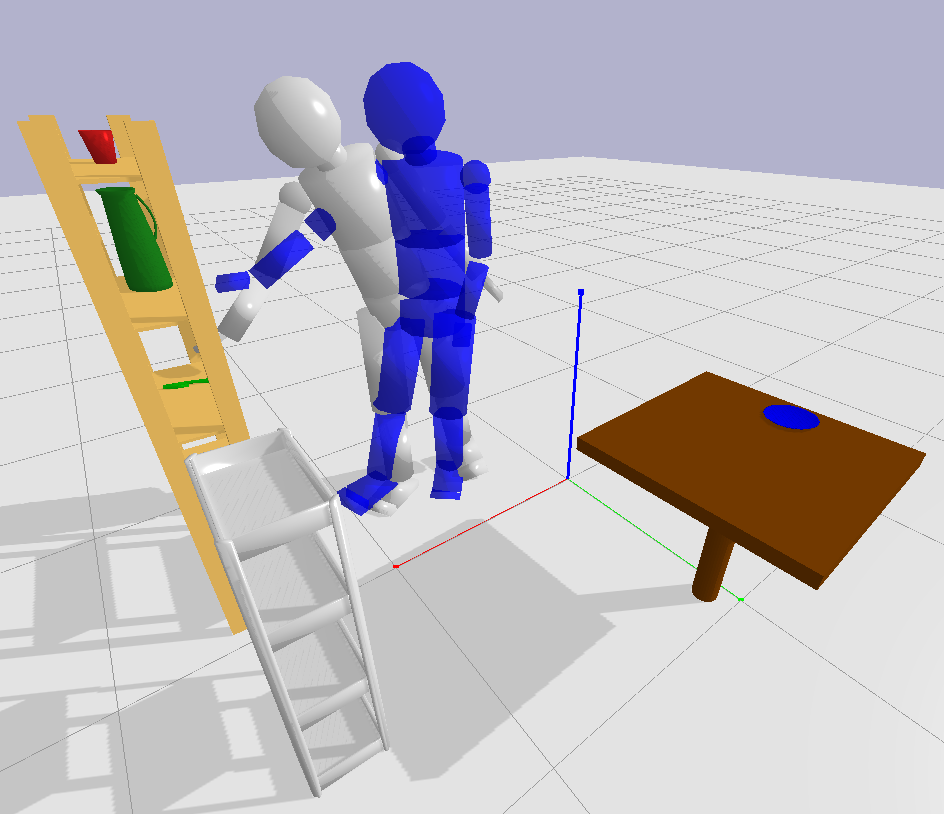}
\end{subfigure}
\\
\centering
\begin{subfigure}{\ltscale\textwidth}
\centering
\includegraphics[width=\linewidth]{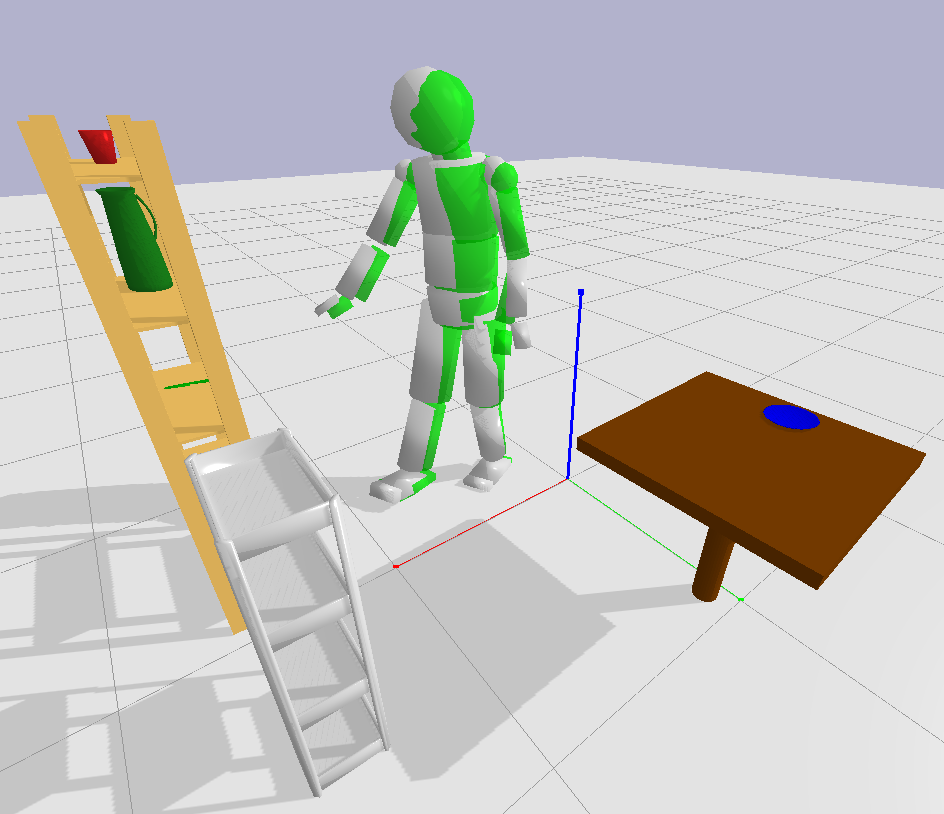}
\end{subfigure}
\hspace{\hsscale}
\begin{subfigure}{\ltscale\textwidth}
\centering
\includegraphics[width=\linewidth]{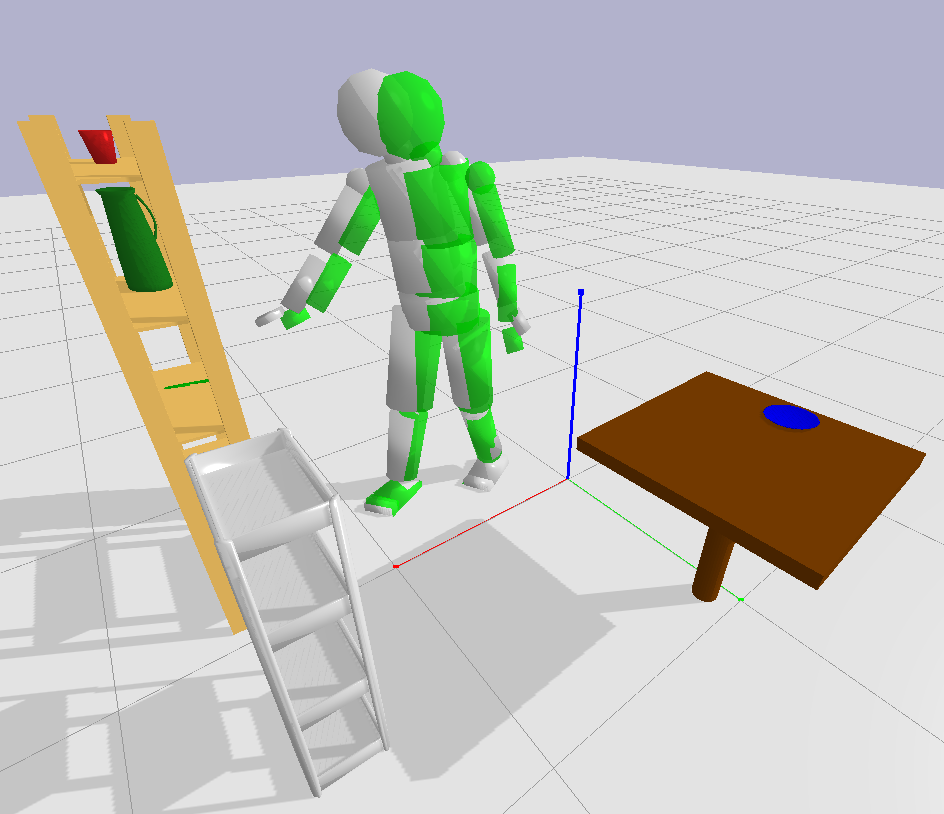}
\end{subfigure}
\hspace{\hsscale}
\begin{subfigure}{\ltscale\textwidth}
\centering
\includegraphics[width=\linewidth]{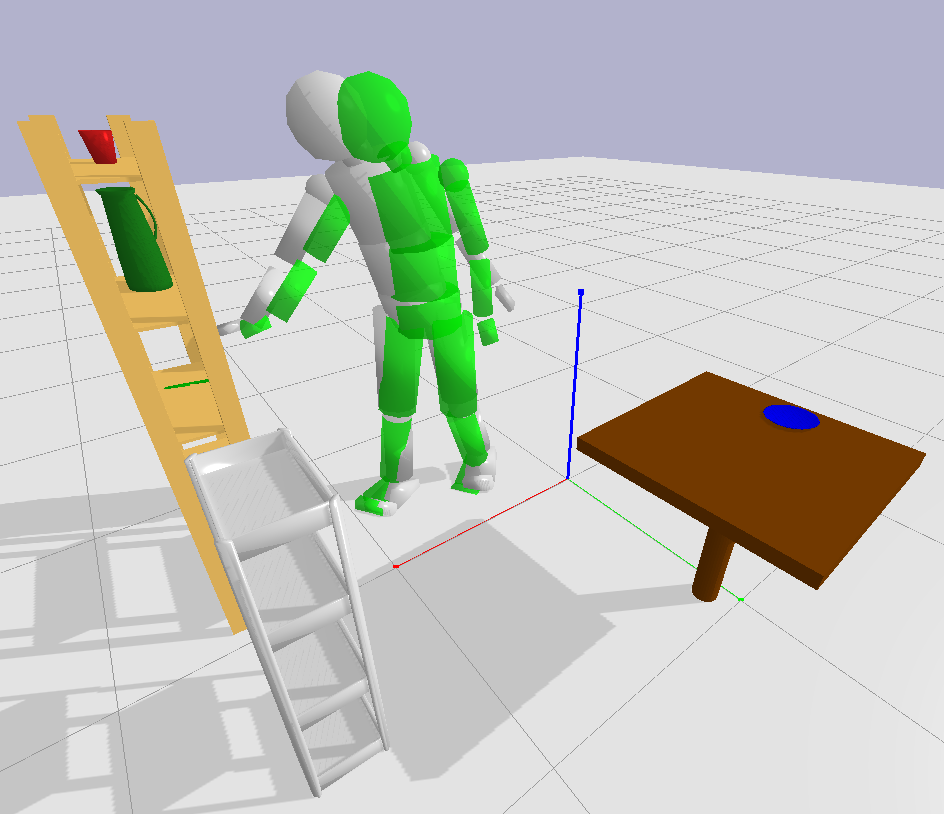}
\end{subfigure}
\hspace{\hsscale}
\begin{subfigure}{\ltscale\textwidth}
\centering
\includegraphics[width=\linewidth]{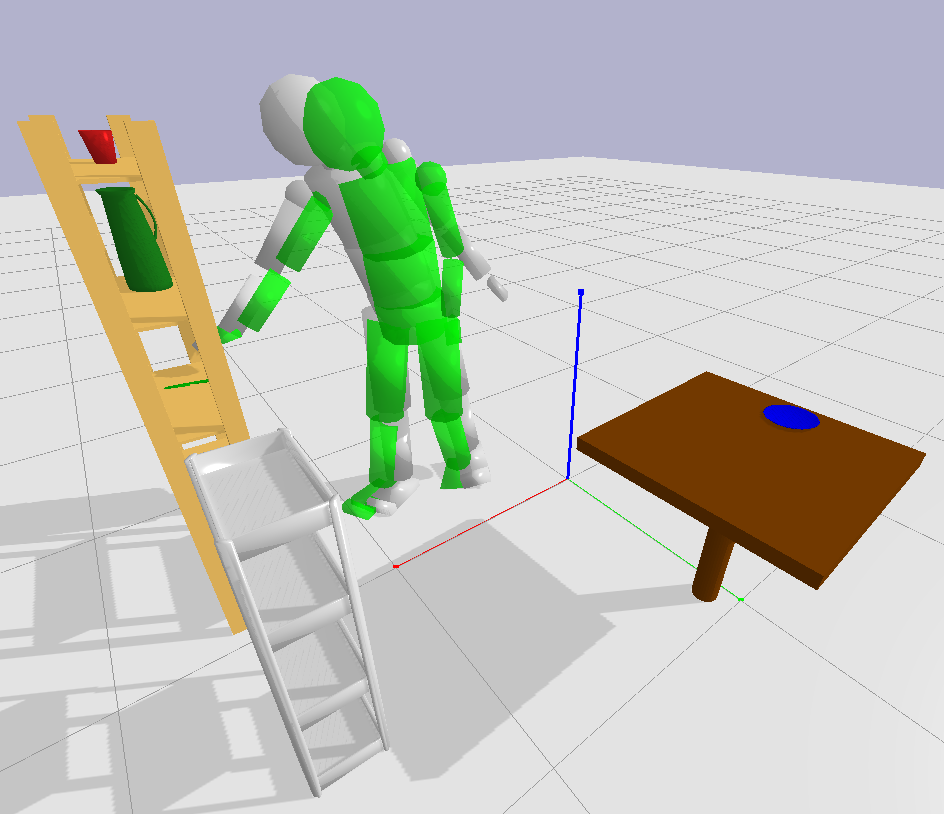}
\end{subfigure}
\hspace{\hsscale}
\begin{subfigure}{\ltscale\textwidth}
\centering
\includegraphics[width=\linewidth]{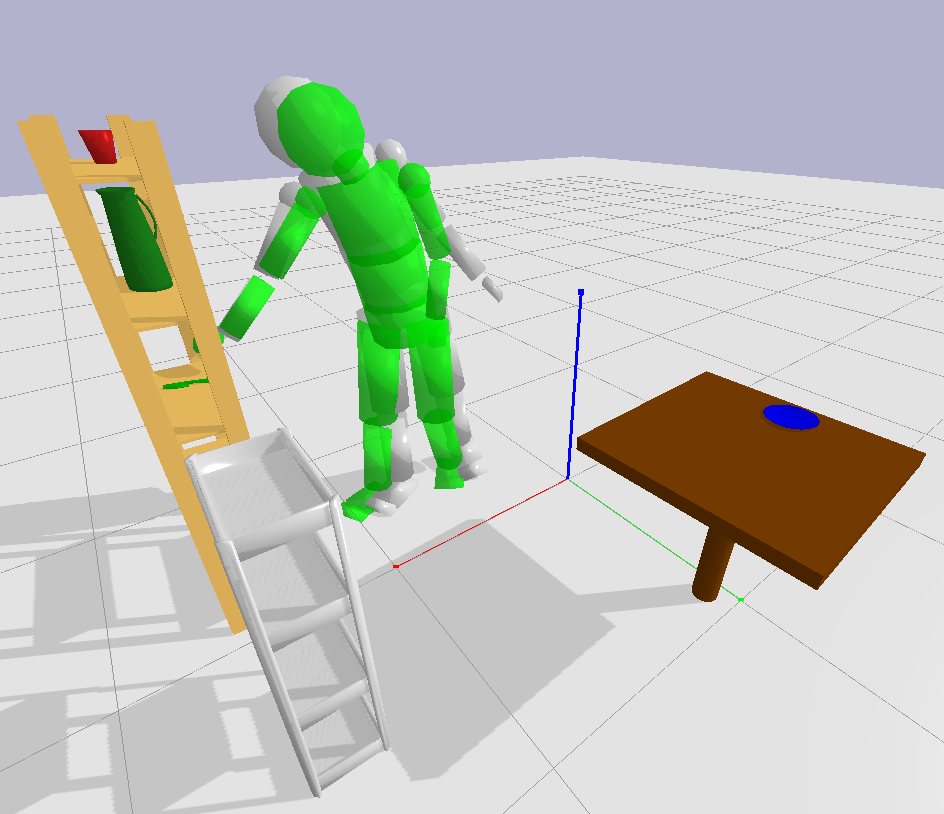}
\end{subfigure}
\caption{Example trajectories. Each row shows one trajectory. From left to right different prediction steps for 0.2 to 1sec in the future are shown. Grey shows the ground truth, green the prediction by our method, blue the prediction without trajectory optimization.}
\label{fig:example_traj}
\vspace{-.2cm}
\end{figure*}

\section{Experiments}
\subsection{Data Set}
The human motion dataset was captured using an Optitrack motion capture system. The subject wore a motion capture suit with 50 markers placed on the full body of the human. The subject was instructed to perform tasks with objects placed on a table, a small shelf and a big shelf in the motion capture area. The used objects were a plate, a knife, a cup, and a jug. The tasks were setting up the table or putting the objects back in the shelves. Marker position data was recorded at a rate of $120$ Hz. In total we recorded 31 minutes of motion capture data. The data includes a set of reaching motions of different objects on different heights.
\subsection{Training the network}
We train our recurrent neural network on 25 minutes of the recorded data set. Fixed length trajectories of 1 second are extracted from the data by using a sliding window. The data is not pre-selected or segmented before training as we want to learn a general dynamic model and not condition on the task of predicting reaching motion. The base transformation of the trajectories is randomized to avoid that the network conditions on real world positions. The network is trained to minimize the mean loss for the prediction of the next 0.5 seconds. Following~\cite{martinez2017human} we use a batch size of 16 and a learning rate of 0.005.

\subsection{Test on reaching trajectories}
\begin{table}
  \centering
\setlength\tabcolsep{5.5pt}
  \begin{tabular}{r|c|c|c|c|c|c|c|c}
  millis & 125 &   250 &   375 &   500 &   625 &   750 &   875 &  1000\\
    \hline\hline
   Zerovel (b) & 0.72 & 1.37 & 1.80 & 2.31 & 2.72 & 3.01 & 3.15 & 3.25\\
    GRU (b) & 0.45 & 0.69 & 0.88 & 1.13 & 1.37 & 1.56 & 1.68 & 1.80\\
    Ours (b) & \textbf{0.42} & \textbf{0.59} & \textbf{0.72} & \textbf{0.87} & \textbf{1.00} & \textbf{1.10} & \textbf{1.15} & \textbf{1.19}\\
   \hline \hline
    Zerovel (w)  & 0.14 & 0.28 & 0.37 & 0.48 & 0.56 & 0.61 & 0.62 & 0.62 \\
    GRU (w) & 0.07 & 0.12 & 0.16 & 0.20 & 0.23 & 0.25 & 0.26 & 0.27 \\
    Ours (w) & 0.06 & \textbf{0.09} & \textbf{0.11} & \textbf{0.12} & \textbf{0.11} & \textbf{0.08} & \textbf{0.05} & \textbf{0.00} \\
    Interp (w) & \textbf{0.05} & 0.11 & 0.15 & 0.17 & 0.17 & 0.13 & 0.08 & \textbf{0.00}\\
  \end{tabular}
  \caption{Error of state prediction on different time steps in the future for the whole body (b). Error of the right wrist (w) for different time steps in the future.}
  \vspace{-.7cm}
  \label{tab:pred_methods}
\end{table}

For testing we use 5 minutes of the recorded data set and extract 25 reaching trajectories from it. We compare our method with three baselines. We compute the distance of key joints of the human (wrists, elbows, knees, ankles and base) and compute the mean distance of the predictions to the ground truth (see Table~\ref{tab:pred_methods}).

The zero velocity baseline predicts the same state for all future steps. The GRU baseline is just the prediction network without information of the goal state. Our method (GRU trajopt) is informed with the goal position of the hand. Table~\ref{tab:pred_methods}~(b) shows the sum of the mean distances of the 6 key joints. It can be seen that the use of trajectory optimization improves the prediction among all future steps.

In Table~\ref{tab:pred_methods}~(w) we only compute the distance of the wrist to the ground truth. We also compute a linear interpolation baseline between the start position of the wrist and the target position. The interpolation baseline and our method are informed with the goal state and thus able to get a zero error in the last time step. Our method also performs better than the interpolation baseline, because the recurrent neural network implicitly reconstructs the underlying human dynamics and not just performs a linear interpolation.

Figure~\ref{fig:example_traj} shows 3 example trajectories from the data with both, the GRU method and our method with added trajectory optimization. It can be seen that the predicted trajectories by our method are close to the ground truth and improve the prediction. Setting the goal position of the wrist clearly helps to reconstruct the full trajectory towards the target, although only the target position of one hand is given.

\section{Conclusions and Future Work}
\balance
We show that a combination of data-driven dynamic models and trajectory optimization for motion prediction works not only with the use of Gaussian processes, as we did in our prior work~\cite{kratzer2018}, but also with the use of a recurrent neural network architecture. In contrast to our GP model we now predict on full-body motion.

Preliminary results on reaching motions demonstrate the efficacy of the approach. The experiments show that the prediction of a state-of-the art recurrent neural network model can be improved by optimizing for an additional goal constraint without changing the training procedure. 

As the neural network is capable of much more training data than the GP without increasing the prediction time, in future work, we will increase the amount of training data and incorporate existing motion data bases. Moreover, we want to add additional optimization objectives, like collision avoidance, to predict trajectories involving obstacles more accurately.
\label{sec:conclusions}

\section*{ACKNOWLEDGMENT}
This work is partially funded by the research alliance ``System Mensch''.
The authors thank the International Max Planck Research School for Intelligent Systems (IMPRS-IS) for supporting Philipp Kratzer.

\pagebreak
\bibliographystyle{plainnat}
\bibliography{references}

\end{document}